\documentclass[journal,twoside,web]{ieeecolor}
\usepackage{generic}
\usepackage{cite}
\usepackage{amsmath,amssymb,amsfonts}
\usepackage{algorithm}
\usepackage{algorithmic}
\usepackage{graphicx}
\usepackage{etoolbox}
\makeatletter
\patchcmd{\pdftex@def@error}
  {\@latex@error}
  {\typeout{WARNING: Missing image file, skip.}\@latex@warning}
  {}{}
\makeatother
\usepackage{caption}
\usepackage{subcaption}
\usepackage{booktabs}
\usepackage{multirow}
\usepackage{array} 
\usepackage{hyperref}
\usepackage{textcomp}
\def\BibTeX{{\rm B\kern-.05em{\sc i\kern-.025em b}\kern-.08em
    T\kern-.1667em\lower.7ex\hbox{E}\kern-.125emX}}
\begin{document}
\title{LGFNet: A CTC-Guided Local–Global Fusion Framework for Single-Channel Sleep Staging}
\author{Chongjian Wang, Zhenghang Hou, Junjie Gao, Xiaofang Zhong, \\
Shiyuan Han, and Tong Zhang,\IEEEmembership{Member, IEEE}
\thanks{Chongjian Wang and Zhenghang Hou are with the School of Mathematics and Systems Science, Shandong University of Science and Technology, Qingdao, 266590, Shandong, China (e-mail: 202311080223@sdust.edu.cn, 202411080404@sdust.edu.cn)}
\thanks{Junjie Gao, Xiaofang Zhong and Shiyuan Han are with the School of Artificial Intelligence, Shandong Women’s University, Jinan, 250352, Shandong, China
(e-mail: junjie.gao@sdwu.edu.cn, 34046@sdwu.edu.cn, ai$\_$hansy@sdwu.edu.cn)}
\thanks{Tong Zhang is with the School of Computer Science and Engineering, South China University of Technology, Guangzhou, 510641, Guangdong, China(e-mail: tony@scut.edu.cn)}
\thanks{Corresponding author: Junjie Gao.}}
\maketitle

\begin{abstract}
Sleep staging remains challenging due to long-range temporal dependencies, ambiguous stage transitions—particularly in N1—and substantial distribution shifts across subjects, sampling rates, and EEG montages. These difficulties are further amplified in single-channel, low-latency scenarios required by wearable and real-world applications. To address these issues, we propose LGFNet, a CTC-guided sequence-to-sequence framework for robust sleep staging. LGFNet introduces a Local–Global Fusion encoder that jointly models fine-grained temporal dynamics and long-range sleep structure, overcoming the limitations of conventional serial hybrid architectures. A CTC–Attention joint training paradigm is adopted to unify temporal alignment with context-dependent modeling, enabling more accurate recognition of stage boundaries and transitions. Furthermore, a three-stage decoding strategy is devised, leveraging CTC-guided decoding and Viterbi-based smoothing to reduce error accumulation and enforce physiological consistency. Extensive cross-dataset evaluations on five public benchmarks demonstrate that LGFNet consistently outperforms state-of-the-art single-channel methods. In particular, on Sleep-EDF-78, LGFNet surpasses DMIN by +1.27\% accuracy, +1.74\% macro-F1, and +1.93\% $\kappa$, with pronounced gains on N1 and transition segments, highlighting its robustness and strong generalization across diverse sampling rates, montages, and recording environments.
\end{abstract}


\section{Introduction}
\IEEEPARstart{S}{leep} staging is fundamental in sleep medicine and wearable health, serving as the basis for diagnosing sleep disorders, assessing sleep quality, and supporting personalized long-term monitoring~[1]. Polysomnography (PSG), which records multiple physiological modalities such as EEG and EOG, remains the clinical gold standard [2], [3]. However, manual scoring of PSG is labor-intensive, subjective, and prone to inconsistencies [4], highlighting the need for accurate and scalable automated sleep staging systems.

Although machine learning and deep learning approaches have achieved promising progress, automated sleep staging remains challenging due to the inherent non-stationarity of EEG signals, susceptibility to artifacts, and substantial inter-individual variability across subjects, sampling rates, and electrode configurations. These factors cause significant distribution shifts that limit the cross-dataset generalization of existing models [5], [6]. Moreover, many methods struggle to capture long-range dependencies and accurately recognize ambiguous transitions—especially the notoriously difficult N1 stage.

In practical clinical and wearable scenarios, constraints such as single-channel acquisition, low latency, and computational efficiency further restrict the applicability of multimodal or resource-intensive decoding frameworks [7], [8]. These constraints call for a sleep staging system that is both lightweight and robust, while still capable of modeling complex temporal patterns inherent in sleep architecture.

To address these challenges, this work introduces LGFNet, a CTC-guided sequence-to-sequence framework specifically designed for single-channel EEG-based sleep staging. LGFNet integrates a Local–Global Fusion encoder that captures fine-grained temporal patterns and long-range sleep structure in parallel, enabling effective modeling of both short- and long-term dependencies. A Transformer-based decoder further captures inter-stage contextual relationships and temporal ordering. Training is conducted under a multi-objective learning scheme that jointly optimizes Connectionist Temporal Classification (CTC) and attention-based cross-entropy [9], [10], yielding robust temporal alignment and context-aware prediction. To ensure physiologically coherent outputs, LGFNet employs a lightweight three-stage decoding strategy that combines CTC prefix-guided decoding, autoregressive refinement, and physiologically informed Viterbi smoothing [11], [12], resulting in an accurate, efficient, and deployment-ready solution for real-world sleep staging.

We comprehensively evaluated LGFNet on five publicly available sleep staging datasets, adopting a strict subject-wise split strategy and fully preserving the original sampling rates and channel configurations of each dataset without any resampling. Experimental results demonstrate that our method achieves stable and significant performance improvements across different sampling rates (100/125/200Hz), different electrode montages (Fpz-Cz vs. C3-A2/M2), and different recording scenarios (clinical vs. in-home). On the Sleep-EDF-78 dataset, LGFNet achieves 88.7\% accuracy (a 1.7\% improvement over the current SOTA method FlexibleSleepNet), 85.0\% macro F1-score (a 2.3\% improvement), and 0.845 $\kappa$ coefficient (a 2.5\% improvement), fully demonstrating its superior robustness. Additionally, on the more challenging ISRUC and SHHS datasets, LGFNet also exhibits outstanding cross-domain generalization capability, achieving substantial performance gains over the existing best single-channel methods.
The major contributions are as follows:

\textbf{\textbullet} A Local–Global Fusion encoder is proposed to jointly model fine-grained temporal features and long-range sleep structure, addressing the limitations of serial hybrid architectures.

\textbf{\textbullet} A CTC–Attention joint training paradigm is introduced to unify temporal alignment and context-dependent modeling, improving recognition of stage boundaries and transitions.

\textbf{\textbullet} A three-stage decoding strategy is devised to reduce error accumulation and enforce physiological consistency through CTC-guided decoding and Viterbi smoothing.

\textbf{\textbullet} A cross-dataset evaluation framework is established, demonstrating the robustness and generalizability of LGFNet across diverse sampling rates, montages, and recording environments.

\section{Related Work}
{\textbf{Machine Learning-Based Methods.}} Automatic sleep staging has historically transitioned from rule-based heuristics to traditional machine-learning pipelines prior to the advent of deep neural architectures [13], [14], [15], [16]. Early rule-based systems, developed in accordance with AASM or R\&K scoring criteria [17], [18], employed expert-specified thresholds on amplitude, spectral power, and band-ratio measurements. While such approaches offered high physiological interpretability, their limited feature expressiveness resulted in modest accuracy and poor robustness to inter-subject variability and noise. Subsequent machine-learning methods sought to mitigate these limitations through handcrafted feature engineering across multiple domains—time-domain statistics and Hjorth parameters, frequency-domain descriptors such as power spectral density and spectral entropy, time–frequency representations  via wavelet transform, and nonlinear dynamical measures—followed by shallow classifiers including SVMs [19] and random forests [20]. Nevertheless, their performance remained tightly coupled to the completeness and discriminability of manually designed features, leaving cross-subject generalization fragile and often insufficient for large-scale or heterogeneous sleep datasets.

{\textbf{Deep Learning–Based Methods.}}
Deep learning has largely eliminated the reliance on handcrafted features by enabling end-to-end representation learning from raw or minimally processed physiological signals. Convolutional architectures (e.g., DeepSleepNet~[21], TinySleepNet~[22], AttnSleep [23], DilatedSleepNet~[24], SleepEEGNet~[25]) substantially improved spectral–temporal feature extraction through architectural refinements, multi-scale filters, and attention mechanisms. Hybrid CNN–RNN models (e.g., MVF-SleepNet~[26], XSleepNet1/2~[27], SeqSleepNet~[28]) further incorporated recurrent units to capture inter-epoch temporal dependencies and regularities in sleep-stage transitions. More recently, self-attention has been leveraged to replace recurrent modules, enabling long-range dependency modeling; Transformer-based designs (e.g., SleepTransformer~[29], Multi-Channel Transformer~[28], SleepViTransformer~[30]) and CNN–Transformer hybrids (e.g., SailentSleepNet~[31]) integrate global context with localized representation learning. In parallel, graph neural networks (e.g., MSTGCN~[32], GraphSleepNet~[33]) explicitly encode spatial relationships in multi-channel EEG to improve robustness across subjects and sensor configurations. Despite these advances, several challenges remain: prevailing models typically optimize a single objective (e.g., cross-entropy or CTC), under-utilizing complementary supervisory signals; temporal inference pipelines often fail to fully exploit physiological transition priors, which are particularly critical for ambiguous stages such as N1; and graph-based or multi-channel frameworks may require extensive preprocessing and introduce substantial computational overhead.

{\textbf{Multimodal Coordination.}}
Multimodal coordination seeks to enhance sleep-stage discrimination—particularly minority stages and artifact resilience—by incorporating auxiliary signals such as EOG, EMG, ECG, respiration, and SpO2 through early fusion, late fusion, or cross-modal attention mechanisms~[28], [34], [35], [36], [37]. However, practical limitations remain prominent: heterogeneous sensors and sampling rates introduce challenges in temporal alignment and calibration~[8], [34], [38]; incomplete modality coverage in many datasets and consumer wearables leads to missing-channel or partially observed modalities~[39]; and multimodal architectures often incur substantial increases in model size, computational load, and susceptibility to cross-device or cross-domain degradation~[39], [40], [41], [42]. Additionally,
multimodal data often requires the placement of different types
of electrodes or sensors on multiple body parts, potentially exacerbating
discomfort during sleep~[7], [8], [43], [44]. In contrast, single-channel EEG (typically frontal or central leads) is widely accessible in home-monitoring and wearable contexts, offers an advantageous signal-to-information balance, and circumvents the modality-availability constraints inherent to multimodal systems~[38], [45]. Consequently, single-channel approaches that effectively encode temporal priors and maximize information extraction from limited inputs are particularly valuable for real-world translation and deployment.

{\textbf{Transformers.}}
Transformers originated in NLP and have since been widely adopted in vision~[46], [47], audio~[48], [49], and multimodal learning~[50], [51], [52] due to their capacity to model long-range dependencies through content-adaptive self-attention. In sleep staging, they enable effective modeling of overnight sequences spanning hundreds of epochs; however, standard Transformer architectures suffer from quadratic time–memory complexity, are prone to overfitting on modest clinical datasets~[50], [51], [52], and generally lack mechanisms for incorporating explicit stage-transition priors. Autoregressive decoders further exhibit “all-SOS” input bias during inference, while combining heterogeneous emission sources (decoder vs. CTC) often introduces scale mismatch~[53]. Although recent studies have begun adapting joint CTC+autoregressive training paradigms from speech recognition (e.g., adaptive channel selection under CTC supervision)~[27], [31], [54], [55], these issues remain only partially resolved. To address them, we designed an LGFM-Encoder that integrates multi-head self-attention for global context with a lightweight convolutional gated-MLP branch for efficient local feature modeling, alongside two prediction heads: a CTC head providing alignment-aware supervision and an autoregressive decoder trained with cross-entropy. During inference, we guide decoder inputs using per-epoch CTC predictions to mitigate SOS-only bias, re-normalize CTC emissions by down-weighting blanks to correct scale mismatch, and fuse the two emission streams—via probability-domain linear fusion or log-domain geometric fusion—before applying Viterbi decoding with data-driven transition priors. This unified training–inference framework combines learnable emissions with physiologically grounded temporal constraints, substantially improves challenging stages such as N1, and remains computationally efficient for long single-channel overnight recordings.

\begin{figure*}[!t]
    \centering

    \includegraphics[width=0.98\textwidth]{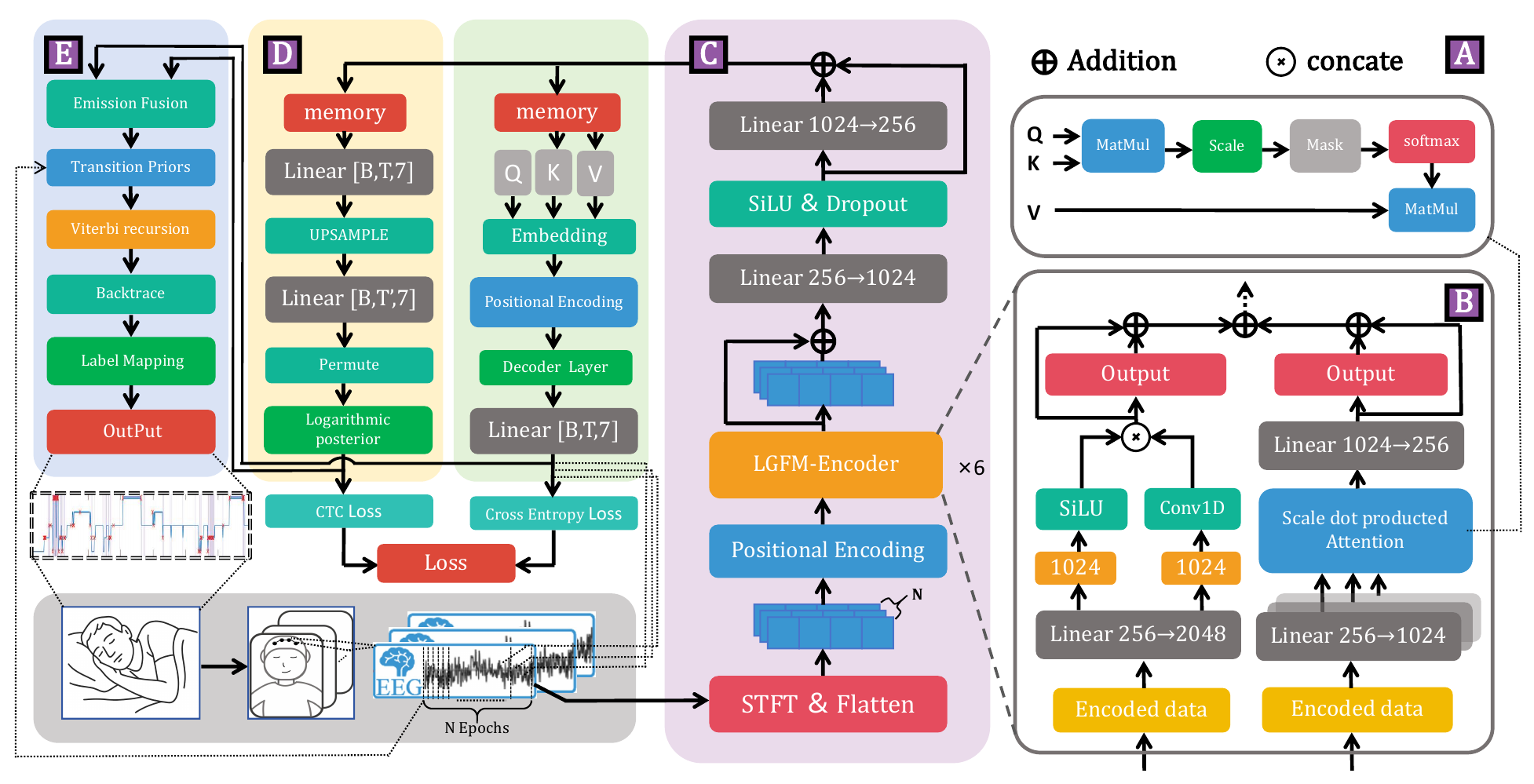}
    \caption{LGFNet pipeline. (A) Apply STFT (2s window, 1s hop) to 30 s EEG epochs.
    (B) Encode with the LGFM-Encoder using parallel LFM (local, gated depthwise-conv MLP)
    and GFM (global, multi-head self-attention) branches.
    (C) Decode with a Transformer via CTC-guided masked autoregression.
    (D) Train with a hybrid CTC/Attention objective (with optional time upsampling and blank suppression).
    (E) At inference, temperature-normalize decoder and CTC emissions, fuse in the probability or log domain,
    and run Viterbi with a five-state transition and initial prior to produce the W/N1/N2/N3/REM sequence.}

    \vspace{0.6em}

    \begin{subfigure}[t]{0.48\textwidth}
        \centering
        \includegraphics[width=\linewidth]{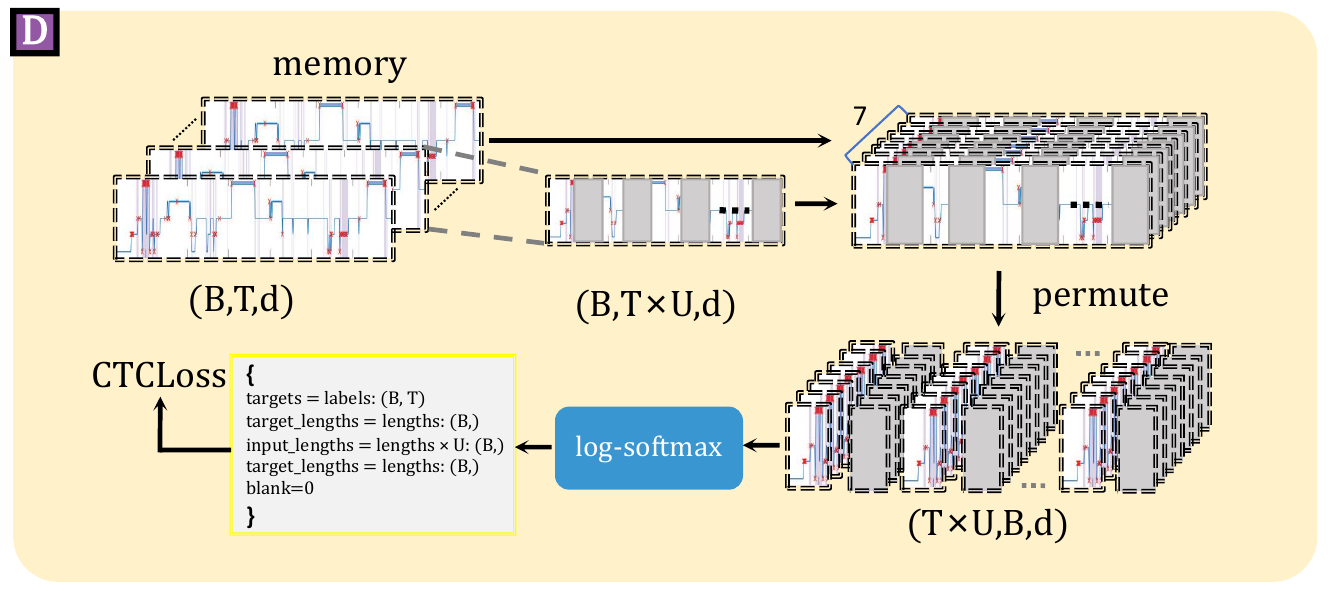}
        \captionsetup{width=\linewidth, justification=raggedright, singlelinecheck=false}
        \caption{\textbf{Figure 2:} CTC-guided training and decoding. During training, the CTC head produces frame-level emissions and pseudo-labels that guide the decoder via scheduled sampling under a hybrid CTC/Attention objective. At inference, CTC-guided masked autoregressive decoding is applied, with optional fusion of decoder and CTC emissions to generate frame-level predictions.}
        \label{fig2}
    \end{subfigure}
    \hfill
    \begin{subfigure}[t]{0.48\textwidth}
        \centering
        \includegraphics[width=\linewidth]{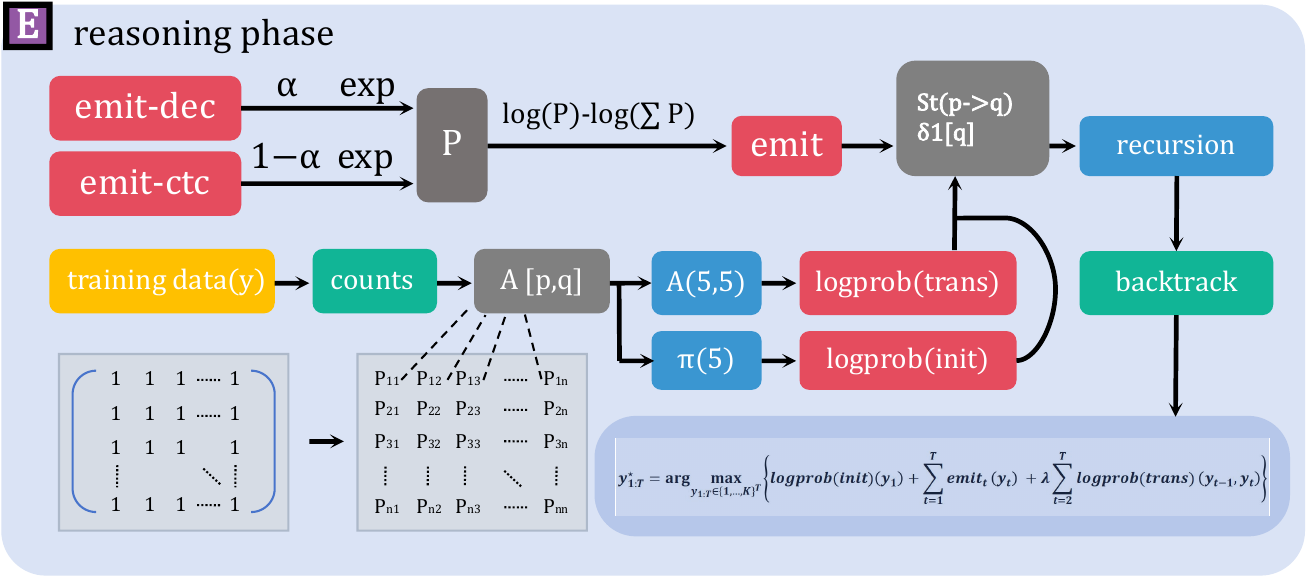}
        \captionsetup{width=\linewidth, justification=raggedright, singlelinecheck=false}
        \caption{\textbf{Figure 3:} Viterbi smoothing. Frame-level emissions from the CTC-guided decoder and the encoder CTC are normalized by temperature and fused in either the probability or log domain, followed by Viterbi decoding with a five-state transition prior and an initial prior estimated from training subjects to produce the W/N1/N2/N3/REM sequence (with an optional argmax mode).}
        \label{fig3}
    \end{subfigure}

    \label{fig1}
\end{figure*}

\section{Method}
\renewcommand{\thesubsection}{\Alph{subsection}}
\setcounter{subsection}{0}   
\subsection{Overview of LGFNet}

As shown in Fig.\ref{fig1}, we propose LGFNet, an end-to-end sequence-to-sequence framework for single-channel EEG sleep staging. The architecture consists of a parallel Local-Global Fusion Multi-Encoder (LGFM-Encoder) and an autoregressive Transformer Decoder. The model is trained under a hybrid CTC/Attention objective, ensuring both alignment and contextual modeling.

Given raw input $X \in \mathbb{R}^{B \times T \times L}$ (batch $B$, sequence length $T$, samples per epoch $L$), we apply a time-parameterized STFT (2s window, 1s hop) to each epoch to obtain time--frequency features $F \in \mathbb{R}^{B \times T \times D_{\mathrm{feat}}}$; this front end preserves native sampling rates without resampling, improving cross-dataset robustness.

The LGFM-Encoder maps the input \( F \) to high-level representations \( H_{\mathrm{enc}} \in \mathbb{R}^{B \times T \times D_{\mathrm{model}}} \). Each encoder layer fuses, in a residual manner, a Local Fusion Module (LFM) and a Global Fusion Module (GFM). Let \( z = \mathrm{LN}(x^{\ell}) \), where \( \mathrm{LN} \) denotes layer normalization. The global branch utilizes multi-head self-attention, defined as

\[
\mathrm{Attn}(Q, K, V) = \mathrm{softmax}\left(\frac{QK^{\top}}{\sqrt{d_k}}\right)V, \quad Q = K = V = z,
\]
yielding \( h_{\mathrm{gfm}} = \mathrm{MHA}(z) \). The local branch employs a gated depthwise-convolutional MLP, expressed as

\[
h_{\mathrm{lfm}} = \sigma(W_g z) \odot \mathrm{DWConv}(W_v z),
\]
which captures fine-grained waveform patterns. The output of the layer is then computed as

\[
x^{\ell+1} = x^{\ell} + \mathrm{Drop}\left( h_{\mathrm{gfm}} + h_{\mathrm{lfm}} \right) + \mathrm{FFN} \left( \mathrm{LN}(x^{\ell}) \right),
\]
thus encoding both short-term details and long-range dependencies at the same depth.

To provide complementary supervision for both alignment and contextual learning, we attach a CTC head to \( H_{\mathrm{enc}} \) and a classification head to the decoder. The CTC branch produces a per-step distribution with a blank, \( p_{\mathrm{ctc}}(\cdot \mid H_{\mathrm{enc}}) \), optionally with \( U \)-fold time upsampling to sharpen the boundaries. The CTC loss is given by

\[
\mathcal{L}_{\mathrm{CTC}} = -\log P_{\mathrm{ctc}}(y \mid H_{\mathrm{enc}}),
\]
where \( P_{\mathrm{ctc}} \) marginalizes over all alignment paths. The decoder, trained with teacher forcing, generates \( P_{\mathrm{att}}(y_t \mid y_{<t}, H_{\mathrm{enc}}) \), and the corresponding loss is
\[
\mathcal{L}_{\mathrm{Att}} = -\sum_{t=1}^{T} \log P_{\mathrm{att}}(y_t \mid y_{<t}, H_{\mathrm{enc}}).
\]

The overall loss is then computed as:
\[
\mathcal{L} = \lambda \,\mathcal{L}_{\mathrm{CTC}} + (1-\lambda)\,\mathcal{L}_{\mathrm{Att}}, \quad \lambda \in [0,1],
\]
where CTC enforces robust frame/segment alignment and boundary sensitivity, while the decoder captures stage order and long-range dependencies, which is particularly beneficial for difficult transitions, such as from N1.

At inference, we support efficient greedy decoding and prior-constrained Viterbi decoding. To prevent SOS-collapse from constant inputs, we first use the CTC framewise argmax to guide the decoder inputs (CTC-guided). We then form the fused emissions as

\[
e_t(c) = (1-\alpha)\,\log P_{\mathrm{att}}(c \mid y_{<t}, H_{\mathrm{enc}}) + \alpha\,\log p_{\mathrm{ctc}}(c \mid H_{\mathrm{enc}}),
\]
and perform Viterbi search with a physiologically informed transition prior \( A \) and initial prior \( \pi \), given by
\[
\hat{y}_{1:T} = \arg\max_{y_{1:T}} \left[ \sum_{t=1}^{T} e_t(y_t) + \beta \log A_{y_{t-1}, y_t} + \log \pi_{y_1} \right],
\]
where \( \alpha \) and \( \beta \) balance emissions and transitions. This procedure ensures low latency and single-channel deployability, while improving temporal consistency and overall accuracy, particularly for transition segments.

\subsection{Time-Frequency Feature Extraction}
Given the non-stationary nature of EEG signals, pure time-domain analysis is insufficient for capturing the dynamic frequency-domain variations associated with sleep stages, such as sleep spindles and K-complexes during N2. To address this challenge, we apply the short-time Fourier transform (STFT) to transform the one-dimensional signal into a two-dimensional time-frequency representation.

Mathematically, for a discrete EEG sequence $x[n]$, the STFT is defined as the windowed Fourier transform with a sliding window:
\begin{equation}
\mathbf{S}(m, k) = \sum_{n} x[n] \, w[n-mR] \, e^{-j \frac{2\pi}{N} k n},
\end{equation}
where $m$ and $k$ denote the time-frame index and frequency-bin index, respectively, $w[n]$ is the window function (a Hann window is adopted in this study to suppress spectral leakage), $N$ is the FFT length, and $R$ is the hop size. 
We compute the magnitude $|\mathbf{S}(m, k)|$ to construct the spectrogram and transpose it to match the model input dimensions. 

This process transforms the raw input $\mathbf{X} \in \mathbb{R}^{B \times T \times L}$ into a time-frequency feature sequence $\mathbf{F}\in\mathbb{R}^{B \times T \times D_{\text{feat}}}$ that is rich in both transient physiological events and background rhythms, thereby providing the encoder with a robust and informative representation.

\subsection{CTC Branch and Training-Time Guidance}
As shown in Fig.2, we employ Connectionist Temporal Classification (CTC) on the encoder outputs to induce robust, monotonic alignment and boundary sensitivity without frame-level supervision. Let the encoder produce $H_{\mathrm{enc}} \in \mathbb{R}^{B\times T\times D}$. To refine temporal resolution near boundaries, we optionally upsample the time axis by a factor $U \in \mathbb{N}$:
\begin{equation}
T' \;=\; U\,T .
\end{equation}
At each step $t \in \{1,\dots,T'\}$, the CTC head emits a distribution over the label set $\mathcal{C}$ augmented with the blank symbol $\varnothing$:
\begin{equation}
p_{\mathrm{ctc}}\!\left(\cdot \mid H_{\mathrm{enc}}, t\right) \in \Delta^{|\mathcal{C}|+1}.
\end{equation}
Given a target sequence $y=(y_1,\dots,y_T)$ and its path set $\mathcal{B}^{-1}(y) \subset (\mathcal{C}\cup\{\varnothing\})^{T'}$ under the collapsing operator $\mathcal{B}$ (removing repeats and blanks), the CTC likelihood and loss are
\begin{equation}
P_{\mathrm{ctc}}(y \mid H_{\mathrm{enc}}) \;=\; \sum_{\pi \in \mathcal{B}^{-1}(y)} \;\prod_{t=1}^{T'} p_{\mathrm{ctc}}\!\left(\pi_t \mid H_{\mathrm{enc}}, t\right),
\end{equation}
\begin{equation}
\mathcal{L}_{\mathrm{CTC}} \;=\; -\log P_{\mathrm{ctc}}(y \mid H_{\mathrm{enc}}),
\end{equation}
computed via the forward--backward dynamic program with log-sum-exp stabilization, which enforces monotonic alignment while accommodating variable segment durations.

To counter early blank dominance and sharpen decision boundaries, we calibrate emissions with temperature scaling and blank suppression. With $\tau_{\mathrm{ctc}}>0$, define the calibrated log-emission
\begin{equation}
\begin{aligned}
\tilde{e}_t(c)
&\;=\; \frac{1}{\tau_{\mathrm{ctc}}}\,\log p_{\mathrm{ctc}}\!\left(c \mid H_{\mathrm{enc}}, t\right) \\
&\qquad +\; \log\!\Big(1 - p_{\mathrm{ctc}}\!\left(\varnothing \mid H_{\mathrm{enc}}, t\right)\Big), 
\qquad c \in \mathcal{C},
\end{aligned}
\end{equation}
and re-normalize within $\mathcal{C}$ as
\begin{equation}
\tilde{p}_{\mathrm{ctc}}\!\left(c \mid t\right) \;=\; 
\frac{\exp\!\big(\tilde{e}_t(c)\big)}{\sum_{c' \in \mathcal{C}} \exp\!\big(\tilde{e}_t(c')\big)}.
\end{equation}
The joint effect of the upsampling factor $U$ and the temperature $\tau_{\mathrm{ctc}}$ increases temporal resolution near putative boundaries and reduces probability dilution induced by blank dominance.

During training, we use CTC to guide the autoregressive decoder inputs to mitigate exposure bias and avoid collapse under constant $\langle \mathrm{sos}\rangle$ inputs. We first derive stepwise pseudo-labels from the calibrated emissions:
\begin{equation}
\tilde{y}_t \;=\; \arg\max_{c \in \mathcal{C}} \; \tilde{p}_{\mathrm{ctc}}\!\left(c \mid t\right).
\end{equation}
We then apply scheduled sampling between teacher forcing and CTC guidance. Let $s_t \sim \mathrm{Bernoulli}(\gamma)$, where $\gamma \in [0,1]$ may be annealed over training; the effective history token consumed by the decoder at step $t$ is
\begin{equation}
y^{\mathrm{guid}}_{t-1} \;=\; s_t \, y_{t-1} \;+\; (1-s_t)\, \tilde{y}_{t-1}.
\end{equation}
The attention decoder models $P_{\mathrm{att}}(y_t \mid y^{\mathrm{guid}}_{<t}, H_{\mathrm{enc}})$ and is trained with cross-entropy:
\begin{equation}
\mathcal{L}_{\mathrm{Att}} \;=\; -\sum_{t=1}^{T} \log P_{\mathrm{att}}\!\left(y_t \mid y^{\mathrm{guid}}_{<t}, H_{\mathrm{enc}}\right).
\end{equation}
The overall loss is then computed as:
\begin{equation}
\mathcal{L} \;=\; \lambda \,\mathcal{L}_{\mathrm{CTC}} \;+\; (1-\lambda)\,\mathcal{L}_{\mathrm{Att}},
\qquad \lambda \in [0,1].
\end{equation}
In this way, the CTC branch supplies strong, monotonic alignment and boundary cues, while the decoder, under CTC guidance, learns context-dependent stage ordering. Inference-time sequence constraints and global decoding are addressed in the next subsection.

\subsection{Inference with Viterbi Smoothing}

As shown in Fig.3, at evaluation time, we decode sleep stages by combining two framewise emission sources---the decoder emission and the CTC emission---and enforcing physiologically plausible transitions via a Viterbi search on a five-state chain. Let $\mathcal{C}=\{\text{W},\text{N1},\text{N2},\text{N3},\text{REM}\}$ with $|\mathcal{C}|=5$. The decoder emission is obtained by running the autoregressive decoder with CTC-guided inputs to avoid $\langle\mathrm{sos}\rangle$-collapse; concretely, we take the framewise CTC argmax $\tilde{y}_{t-1}$ as the previous token and produce $P_{\mathrm{att}}(\cdot \mid y^{\mathrm{guid}}_{<t}, H_{\mathrm{enc}})$. With a temperature $\tau_{\mathrm{att}}>0$, we form normalized log-emissions
\begin{equation}
\begin{aligned}
e^{\mathrm{dec}}_t(c)
  = &\log\!\Big(P_{\mathrm{att}}(c \mid y^{\mathrm{guid}}_{<t}, H_{\mathrm{enc}})\Big)^{1/\tau_{\mathrm{att}}}\\[0.8ex]
    &- \log \sum_{c'\in\mathcal{C}} \Big(P_{\mathrm{att}}(c' \mid y^{\mathrm{guid}}_{<t}, H_{\mathrm{enc}})\Big)^{1/\tau_{\mathrm{att}}},\\
    &\qquad c\in\mathcal{C}.
\end{aligned}
\end{equation}
which corresponds to temperature scaling in log-space followed by within-class normalization.

The CTC emission is derived from the encoder CTC head while suppressing blank and optionally aggregating $U$-fold upsampled sub-steps. Let $p_{\mathrm{ctc}}(\cdot \mid H_{\mathrm{enc}}, t)$ denote the softmax over $\{\varnothing\}\cup\mathcal{C}$ at step $t$. We first calibrate and suppress the blank posterior and then normalize across $\mathcal{C}$:

\begin{equation}
\begin{aligned}
\tilde{e}^{\mathrm{ctc}}_t(c)
  &= \frac{1}{\tau_{\mathrm{ctc}}}\,\log p_{\mathrm{ctc}}\!\big(c \mid H_{\mathrm{enc}}, t\big)\\[0.8ex]
  &\quad + \log\!\Big(1 - p_{\mathrm{ctc}}\!\big(\varnothing \mid H_{\mathrm{enc}}, t\big)\Big),\\[0.8ex]
e^{\mathrm{ctc}}_t(c)
  &= \tilde{e}^{\mathrm{ctc}}_t(c) \;-\; \log \!\sum_{c'\in\mathcal{C}} \exp\!\big(\tilde{e}^{\mathrm{ctc}}_t(c')\big),
  \qquad c \in \mathcal{C}.
\end{aligned}
\end{equation}
When upsampling is enabled, the encoder step $t$ expands to sub-steps $u=1,\dots,U$. Let $\tilde{e}^{\mathrm{ctc}}_{t,u}(c)$ be the calibrated log-emission at sub-step $u$; we aggregate by log-mean-exp to preserve the log-probability scale,
\begin{equation}
\begin{aligned}
\bar{e}^{\mathrm{ctc}}_t(c)
&\;=\; \log\!\left(\frac{1}{U}\sum_{u=1}^{U}\exp\big(\tilde{e}^{\mathrm{ctc}}_{t,u}(c)\big)\right), \\
e^{\mathrm{ctc}}_t(c)
&\;=\; \bar{e}^{\mathrm{ctc}}_t(c) \;-\; \log \!\sum_{c'\in\mathcal{C}} \exp\!\big(\bar{e}^{\mathrm{ctc}}_t(c')\big),
\qquad c \in \mathcal{C},
\end{aligned}
\end{equation}
which matches the implementation that repeats encoder frames by a factor $U$ and then averages in log-space.

We then fuse the two emission streams into a single normalized log-emission $e_t(c)$ used by Viterbi. Two fusion rules are supported and both return properly normalized log-scores. In the probability-domain mixture, we compute

\begin{equation}
\begingroup
\renewcommand{\arraystretch}{1.35} 
\begin{array}{@{}l@{}}
p^{\mathrm{dec}}_t(c) = \exp\!\big(e^{\mathrm{dec}}_t(c)\big),\\[0.8ex]
p^{\mathrm{ctc}}_t(c) = \exp\!\big(e^{\mathrm{ctc}}_t(c)\big), \qquad c\in\mathcal{C},\\[0.8ex]
p^{\mathrm{fuse}}_t(c) = (1-\alpha)\, p^{\mathrm{dec}}_t(c) + \alpha \, p^{\mathrm{ctc}}_t(c),\\[0.8ex]
e_t(c) = \log \frac{p^{\mathrm{fuse}}_t(c)}{\sum_{c'\in\mathcal{C}} p^{\mathrm{fuse}}_t(c')}.
\end{array}
\endgroup
\end{equation}
where $\alpha \in [0,1]$ controls the trade-off. In the log-domain convex fusion, we directly combine log-emissions and re-normalize,
\begin{equation}
\begin{aligned}
\tilde{e}_t(c) \;=\; (1-\alpha)\, e^{\mathrm{dec}}_t(c) \;+\; \alpha \, e^{\mathrm{ctc}}_t(c), \\
e_t(c) \;=\; \tilde{e}_t(c) \;-\; \log \!\sum_{c'\in\mathcal{C}} \exp\!\big(\tilde{e}_t(c')\big).
\end{aligned}
\end{equation}
Both fusion modes correspond to the implementation switches \texttt{fuse\_rule=prob} and \texttt{fuse\_rule=log}, respectively; temperatures $\tau_{\mathrm{att}}$ and $\tau_{\mathrm{ctc}}$ and the mixing weight $\alpha$ are hyperparameters set on validation.

To impose physiologically informed dynamics, we estimate a first-order transition prior from the training subjects. Let $A \in \mathbb{R}^{5\times 5}$ be the row-stochastic transition matrix with Laplace smoothing, $A_{p,q} \propto \text{count}(y_{t-1}=p,y_t=q)+1$, and let $\pi \in \Delta^{5}$ be the initial-state prior estimated from the first epoch of each recording (or uniform if specified). Given the fused emissions $e_t(\cdot)$, the most probable stage sequence $\hat{y}_{1:T}$ is obtained by Viterbi decoding on the five-state chain with a transition weight $\beta \ge 0$:
\begin{equation}
\begin{aligned}
\text{}\quad & \mathrm{DP}_1(q) \;=\; e_1(q) \;+\; \log \pi_q, \qquad q \in \mathcal{C}, \\
\text{}\quad & \mathrm{DP}_t(q) \;=\; e_t(q) \;+\; \max_{p \in \mathcal{C}} \Big\{ \mathrm{DP}_{t-1}(p) \;+\; \beta \,\log A_{p,q} \Big\}, \\
\text{}\quad & \hat{y}_T \;=\; \arg\max_{q \in \mathcal{C}} \mathrm{DP}_T(q), \quad\\
&\hat{y}_{t-1} \;=\; \arg\max_{p \in \mathcal{C}} \Big\{ \mathrm{DP}_{t-1}(p) \;+\; \beta \,\log A_{p,\hat{y}_t} \Big\}.
\end{aligned}
\end{equation}

This dynamic program has complexity $\mathcal{O}(T K^2)$ with $K=5$ and is thus negligible in practice. For debugging, we also support a degenerate mode that skips Viterbi and outputs $\hat{y}_t=\arg\max_{c} e_t(c)$ directly.

Finally, the decoded path $\hat{y}_{1:T}$ is mapped to the five-way label space $\mathcal{C}$ to realize the five-class staging task. In implementation, states are represented as integers in $\{0,\dots,4\}$ for $\{\text{W},\text{N1},\text{N2},\text{N3},\text{REM}\}$; for compatibility with training targets that reserve \textsc{Pad} and \textsc{Sos} indices, we shift to $\{1,\dots,5\}$ where necessary. All reported metrics (overall accuracy, macro-F1, sensitivity, specificity, and Cohen's $\kappa$) are computed on the fixed five-class space after this mapping.

\begin{table*}[t]
\centering
\caption{Statistics of sleep stage distribution in different datasets}
\label{tab:1}
\small
\setlength{\tabcolsep}{9pt}   
\begin{tabular}{@{}l c c c | c c c c c | r @{}}
\toprule
\multirow{2}{*}{\textbf{Dataset}} 
 & \multirow{2}{*}{\textbf{Subjects}} 
 & \multirow{2}{*}{\textbf{Channel}} 
 & \multirow{2}{*}{\textbf{Freq.}} 
 & \multicolumn{5}{c|}{\textbf{Epoch numbers}} 
 & \multirow{2}{*}{\textbf{Total}} \\
\cmidrule(lr{0.5em}){5-9}
 &  &  &  & Wake    & N1     & N2      & N3     & REM    &  \\
\midrule
Sleepedf-20 & 20  & Fpz-Cz & 100 Hz & 8285  & 2804  & 17799 & 5703 & 7717 & 42308 \\
            &     &        &        & 19.6\% & 6.6\% & 42.1\% & 13.5\% & 18.2\% &       \\
\midrule
Sleepedf-78 & 78  & Fpz-Cz & 100 Hz & 63802 & 21229 & 68645 & 12883 & 25651 & 192210 \\
            &     &        &        & 33.2\% & 11.0\% & 35.71\% & 6.7\% & 13.34\% &       \\
\midrule
SHHS        & 329 & C4-A1  & 125 Hz & 46369 & 10304 & 142125 & 60153 & 65953 & 324854 \\
            &     &        &        & 14.3\% & 3.2\% & 43.7\% & 18.5\% & 20.3\% &       \\
\midrule
ISRUC-S3    & 10  & C4-A1  & 200 Hz & 1651  & 1215  & 2609  & 2014  & 1060  & 8549  \\
            &     &        &        & 20.5\% & 14.2\% & 30.5\% & 23.6\% & 12.4\% &       \\
\midrule
ISRUC-S1    & 100 & C4-A1  & 200 Hz & 20098 & 11062 & 27511 & 17251 & 11265 & 87187 \\
            &     &        &        & 23.1\% & 13.31\% & 31.2\% & 19.8\% & 12.9\% &       \\
\bottomrule
\setlength{\textfloatsep}{6pt}
\setlength{\intextsep}{6pt}
\end{tabular}
\end{table*}

\begin{table}[t]
\centering
\caption{Default hyperparameter settings}
\label{tab:2}
\small
\setlength{\tabcolsep}{9pt}   
\begin{tabular}{@{}l >{\raggedleft\arraybackslash}p{4.5cm}@{}}
\toprule
\textbf{Parameter}              & \textbf{Value}                  \\
\midrule
Core Dimension                  & 256                             \\
Encoder Layers                  & 6                              \\
Attention Heads                 & 8                              \\
Dropout Rate                    & 0.1                            \\
Batch Size                      & 4                              \\
Learning Rate                   & $1.0 \times 10^{-4}$           \\
Optimizer                       & AdamW                          \\
Number of Epochs                & 50                             \\
CTC Loss Weight $\lambda$       & 0.3                            \\
MHA Loss Weight $w$             & 0.3                            \\
Smoothing Factor $\alpha$       & 0.3                            \\
\bottomrule
\end{tabular}
\end{table}

\begin{table*}[!htbp]
\centering
\caption{Comparison with state-of-the-art methods on five public datasets}
\label{tab:3}
\small
\setlength{\aboverulesep}{0.0ex}
\setlength{\belowrulesep}{0.1ex}

\setlength{\tabcolsep}{4.0pt}
\begin{tabular}{@{}l l ccc ccccc c@{}}
\toprule
\textbf{Dataset} & \textbf{Method} & \textbf{Architecture} & \textbf{Acc} & \textbf{MF1} & \textbf{Kappa} & \textbf{W} & \textbf{N1} & \textbf{N2} & \textbf{N3} & \textbf{R} \\
\midrule         
\multirow{11}{*}{ISRUC-S3}
 & MVF-SleepNet~[26]      & CNN + LSTM        & \underline{84.1} & \underline{82.8} & \underline{0.795} & 90.0  & 62.5 & \underline{83.3} & 91.1 & \underline{87.3} \\
 & MSTGCN~[32]             & GCN               & 82.1 & 80.8 & 0.769 & 89.4  & 59.6 & 80.6 & 89.0 & 85.6 \\
 & SVM~[19]              & Traditional       & 71.4 & 67.2 & 0.626 & 82.4  & 42.8 & 72.4 & 81.5 & 56.9 \\
 & RF~[20]                 & Traditional       & 70.2 & 68.5 & 0.616 & 83.8  & 47.0 & 67.1 & 76.3 & 68.4 \\
 & MixSleepNet~[54]        & CNN + Transformer & 83.0 & 82.1 & 0.782 & 89.9  & 62.5 & 81.9 & 89.9 & 86.0 \\
 & XSleepNet1~[27]        & CNN + LSTM        & 82.5 & 80.8 & 0.774 & \underline{90.1}  & 58.6 & 82.5 & 88.7 & 84.3 \\
 & XSleepNet2~[27]         & CNN + LSTM        & 82.6 & 81.0 & 0.774 & 89.9  & 59.0 & 82.6 & 88.4 & 84.9 \\
 & SeqSleepNet~[28]        & LSTM              & 78.9 & 76.3 & 0.730 & 83.6  & 43.9 & 79.3 & 87.9 & 86.7 \\
 & SleepAC~[56]             & Adap-Sel          & \underline{84.1} & 82.4 & 0.788 & 89.7  & \textbf{64.7} & 82.3 & \textbf{89.8} & 85.3 \\
 & \textbf{ours}     & \textbf{CNN + Transformer}    & \textbf{86.1} & \textbf{84.4} & \textbf{0.822} & \textbf{90.3} & \underline{64.5} & \textbf{89.9} & \textbf{86.7} & \textbf{90.4} \\
\midrule
\multirow{11}{*}{ISRUC-S1}
 & MVF-SleepNet~[26]      & CNN + LSTM        & 82.1 & \underline{80.2} & 0.768 & \underline{90.8}  & \underline{56.2} & \underline{81.1} & 87.1 & \underline{85.7} \\
 & MSTGCN~[32]            & GCN               & 80.9 & 78.7 & 0.752 & 89.3  & 53.1 & 79.9 & 86.7 & 84.4 \\
 & SVM~[19]                  & Traditional       & 68.4 & 60.8 & 0.583 & 79.3  & 24.2 & 70.8 & 80.8 & 49.0 \\
 & RF~[20]                 & Traditional       & 69.9 & 64.9 & 0.607 & 84.1  & 30.7 & 70.5 & 75.0 & 64.0 \\
 & MixSleepNet~[54]        & CNN + Transformer & 81.3 & 78.7 & 0.757 & \underline{90.8}  & 51.2 & 79.9 & 87.1 & 84.4 \\
 & XSleepNet1~[27]       & CNN + LSTM        & 80.6 & 78.2 & 0.748 & 89.4  & 50.4 & 80.0 & 87.1 & 83.9 \\
 & XSleepNet2~[27]        & CNN + LSTM        & 80.5 & 78.4 & 0.751 & 89.4  & 50.6 & 80.2 & \underline{87.5} & 84.4 \\
 & SeqSleepNet~[28]       & LSTM              & 77.0 & 68.3 & 0.648 & 84.4  & 12.4 & 76.9 & 85.3 & 79.4 \\
 & SleepAC~[56]             & Adap-Sel          & \underline{82.3} & 80.1 & \underline{0.771} & 90.7  & 55.6 & 81.0 & 87.1 & 85.6 \\
 & \textbf{ours}     & \textbf{CNN + Transformer}    & \textbf{85.7} & \textbf{83.5} & \textbf{0.814} & \textbf{91.3} & \textbf{61.9} & \textbf{86.9} & \textbf{87.8} & \textbf{89.6} \\
\midrule
\multirow{15}{*}{SleepEDF-20}
 & XSleepNet1~[27]        & CNN + LSTM        & 86.0 & 80.0 & 0.810 & 91.3  & 49.5 & 88.0 & 86.9 & 84.2 \\
 & XSleepNet2~[27]        & CNN + LSTM        & 86.3 & 80.6 & 0.813 & 92.2  & 51.8 & 88.0 & 86.8 & 83.9 \\
 & SleepAC~[56]              & Adap-Sel          & 86.3 & 81.2 & 0.801 & 89.7  & 54.0 & 88.3 & 89.0 & 84.9 \\
 & SalientSleepNet~[31]   & CNN-Transformer   & 87.5 & \underline{83.0} & -     & 92.3  & 56.2 & 89.9 & 87.2 & \underline{89.2} \\
 & SeqSleepNet~[28]        & LSTM              & 86.0 & 79.7 & 0.810 & -     & -    & -    & -    & -    \\
 & DilatedSleepNet~[24]   & CNN               & 86.8 & 81.9 & 0.820 & 90.8  & 53.3 & 89.4 & \underline{89.9} & 85.8 \\
 & AttnSleep~[23]         & CNN               & 84.4 & 78.1 & 0.790 & 89.7  & 42.6 & 88.8 & 90.2 & 79.0 \\
 & DeepSleepNet~[21]      & CNN               & 82.0 & 76.9 & 0.760 & 84.7  & 46.6 & 85.9 & 84.8 & 82.4 \\
 & SleepEEGNet~[25]       & CNN               & 84.3 & 79.7 & 0.790 & 89.2  & 52.2 & 86.8 & 85.1 & 85.0 \\
 & FlexibleSleepNet~[55]  & CNN               & 86.9 & 81.9 & 0.824 & 92.8  & \underline{57.2} & 89.8 & 85.0 & 87.2 \\
 & Multi-channel~[57]     & Transformer       & 87.2 & 81.2 & 0.820 & 92.8  & 49.1 & 90.0 & 89.3 & 84.8 \\
 & TinySleepNet~[22]      & CNN               & 85.4 & 80.5 & -     & 90.1  & 51.4 & 88.5 & 88.3 & 84.3 \\
 & SleepViTransformer~[30]& Transformer      & \underline{87.8} & 81.5 & \underline{0.834} & \textbf{93.8}  & 48.4 & 89.2 & 88.4 & 87.9 \\
 & \textbf{ours}     & \textbf{CNN + Transformer}    & \textbf{89.7} & \textbf{85.9} & \textbf{0.859} & \underline{93.7} & \textbf{65.0} & \textbf{90.9} & \textbf{85.5} & \textbf{94.6} \\
\midrule
\multirow{13}{*}{SleepEDF-78}
 & SalientSleepNet~[31]    & CNN-Transformer   & 84.1 & 79.5 & -     & 93.3  & 54.2 & 85.8 & 78.3 & 85.8 \\
 & SeqSleepNet~[28]      & LSTM              & 83.8 & 78.2 & 0.780 & -     & -    & -    & -    & -    \\
 & DilatedSleepNet~[24]  & CNN               & 83.2 & 77.5 & 0.770 & 92.7  & 47.2 & 85.5 & 81.8 & 80.5 \\
 & AttnSleep~[23]        & CNN               & 81.3 & 75.1 & 0.740 & 92.0  & 42.0 & 85.0 & \underline{82.1} & 74.2 \\
 & DeepSleepNet~[21]      & CNN               & 76.9 & 70.7 & 0.690 & 90.8  & 44.8 & 78.5 & 67.9 & 71.3 \\
 & SleepEEGNet~[25]       & CNN               & 80.0 & 73.6 & 0.730 & 91.7  & 44.1 & 82.5 & 73.5 & 75.2 \\
 & SleepTransformer~[29]  & Transformer       & 81.4 & 74.3 & 0.743 & 91.7  & 40.4 & 84.3 & 77.9 & 77.2 \\
 & FlexibleSleepNet~[55]  & CNN               & \underline{87.0} & \underline{82.7} & \underline{0.820} & \textbf{95.3}  & \underline{59.9} & \underline{88.0} & \textbf{84.4} & 86.1 \\
 & Multi-channel~[57]     & Transformer       & 85.0 & 79.6 & 0.790 & \underline{94.0}  & 53.0 & 86.9 & 81.8 & 82.6 \\
 & TinySleepNet~[22]       & CNN               & 83.1 & 78.1 & -     & 92.8  & 51.0 & 85.3 & 81.1 & 80.3 \\
 & SleepViTransformer~[30]& Transformer      & 85.0 & 79.1 & 0.792 & 93.6  & 49.4 & 86.4 & 79.3 & \underline{86.9} \\
  
 & \textbf{ours}     & \textbf{CNN + Transformer}    & \textbf{88.7} & \textbf{85.0} & \textbf{0.845} & 93.7 & \textbf{67.1} & \textbf{90.8} & 80.5 & \textbf{92.8} \\

\midrule
\multirow{11}{*}{SHHS}
 & XSleepNet1~[27]        & CNN + LSTM        & 87.5 & 81.0 & 0.826 & 91.6  & \underline{51.4} & 88.5 & 85.0 & 88.4 \\
 & XSleepNet2~[27]        & CNN + LSTM        & 87.6 & 80.7 & 0.826 & 92.0  & 49.9 & 88.3 & 85.0 & 88.2 \\
 & DilatedSleepNet~[24]    & CNN               & 85.4 & 78.7 & 0.800 & 86.1  & 47.7 & 87.1 & 84.7 & 87.8 \\
 & AttnSleep~[23]         & CNN               & 84.2 & 75.3 & 0.780 & 86.7  & 33.2 & 87.1 & 87.1 & 82.1 \\
 & DeepSleepNet~[21]       & CNN               & 81.0 & 73.9 & 0.730 & 85.4  & 40.5 & 82.5 & 79.3 & 81.9 \\
 & SleepEEGNet~[25]        & CNN               & 73.9 & 68.4 & 0.650 & 81.3  & 34.4 & 73.4 & 75.9 & 77.0 \\
 & SleepTransformer~[29]  & Transformer       & 87.7 & 80.1 & 0.828 & 92.2  & 46.1 & 88.3 & 85.2 & 88.6 \\
 & FlexibleSleepNet~[55]  & CNN               & 87.6 & 79.5 & 0.830 & 92.3  & 40.0 & \underline{88.8} & 87.0 & 89.7 \\
 & Multi-channel~[57]    & Transformer       & 87.5 & 79.3 & 0.820 & 92.3  & 40.5 & 88.6 & 88.2 & 87.7 \\
 & SleepViTransformer~[30]& Transformer      & \underline{88.1} & 79.8 & 0.830 & \textbf{93.4}  & 44.4 & 88.5 & \underline{88.5} & 88.3 \\
 & SleepAC~[56]           & Adap-Sel          & 87.8 & \underline{81.3} & \underline{0.832} & \underline{92.4}  & 51.3 & 88.7 & 84.5 & \underline{89.6} \\
 & \textbf{ours}     & \textbf{CNN + Transformer}    & \textbf{89.9} & \textbf{84.6} & \textbf{0.858} & 91.6 & \textbf{59.3} & \textbf{90.9} & \textbf{89.6} & \textbf{91.7} \\
\bottomrule
\end{tabular}
\end{table*}

\section{Experiment}
\subsection{Datasets}
\label{sec:datasets}

To comprehensively evaluate the performance and generalization capability of LGFNet, we conducted extensive experiments on five publicly available sleep staging benchmarks.The detailed statistics and class distributions of all datasets are summarized in Tab.~\ref{tab:1}.

To ensure fair and consistent comparison across different datasets, all experiments are conducted with a unified set of hyperparameters, as summarized in Tab.~\ref{tab:2}. A strict subject-wise cross-validation strategy was adopted for all datasets, ensuring that the model is evaluated on entirely unseen subjects and thus genuinely reflecting its cross-subject generalization ability. Based on common practice in this field, we use 20-fold cross-validation for Sleep-EDF-20, 10-fold cross-validation for Sleep-EDF-78 and ISRUC S1/S3, and 5-fold cross-validation for SHHS.

Crucially, we deliberately preserved the original sampling rates and electrode montages of each dataset without any resampling or cross-dataset normalization. 
This protocol is designed to emulate the substantial data heterogeneity encountered in real-world clinical and wearable scenarios, thereby providing a more challenging and clinically meaningful assessment of model robustness.

\subsection{Evaluation Metrics}
\label{sec:metrics}

To comprehensively quantify the classification performance and sequence consistency of LGFNet, we adopt the following standard evaluation metrics: overall accuracy (ACC), macro-averaged precision (MPR), macro-averaged recall (MRE), macro-averaged F1-score (Macro F1), and Cohen's kappa coefficient ($\kappa$).

The overall accuracy is defined as
\begin{equation}
\text{ACC} = \frac{\text{TP} + \text{TN}}{\text{TP} + \text{FN} + \text{FP} + \text{TN}},
\label{eq:acc}
\end{equation}
where TP, TN, FP, and FN denote true positives, true negatives, false positives, and false negatives, respectively.

For each class $k$, we compute the per-class precision and recall as
\begin{equation}
\text{PR}_k = \frac{\text{TP}_k}{\text{TP}_k + \text{FP}_k}, \quad
\text{RE}_k = \frac{\text{TP}_k}{\text{TP}_k + \text{FN}_k}.
\end{equation}

The macro-averaged precision and recall are obtained by averaging over all classes:
\begin{equation}
\text{MPR} = \frac{1}{N_{\text{classes}}} \sum_{k=1}^{N_{\text{classes}}} \text{PR}_k,
\end{equation}
\begin{equation}
\text{MRE} = \frac{1}{N_{\text{classes}}} \sum_{k=1}^{N_{\text{classes}}} \text{RE}_k.
\end{equation}

The macro-averaged F1-score (Macro F1) harmonically combines macro precision and recall:
\begin{equation}
\text{Macro F1} = \frac{2 \times \text{MPR} \times \text{MRE}}{\text{MPR} + \text{MRE}}.
\label{eq:macrof1}
\end{equation}

Cohen's kappa coefficient ($\kappa$) measures the agreement between the predicted and true labels beyond chance, which is particularly suitable for imbalanced classification tasks:
\begin{equation}
\kappa = 1 - \frac{1 - \text{ACC}}{1 - p_e},
\end{equation}
where ACC is defined in Eq.~\eqref{eq:acc}, and $p_e$ is the probability of agreement by chance, calculated as
\begin{equation}
p_e = \frac{1}{N^2} \sum_{k=1}^{N_{\text{classes}}} n_{k1} n_{k2},
\end{equation}
with $N$ being the total number of samples, $n_{k1}$ the number of samples truly belonging to class $k$, and $n_{k2}$ the number of samples predicted as class $k$.

\begin{table}[t]
\caption{Sleep stage classification performance (\%)}
\centering
\label{tab:4}
\footnotesize
\setlength{\tabcolsep}{5pt}
\begin{tabular}{@{}llcccc>{\raggedleft\arraybackslash}p{1.9cm}@{}}
\toprule
\textbf{Dataset} & \textbf{Stage} & \textbf{Prec.} & \textbf{F1} & \textbf{Spec.} & \textbf{Support} \\
\midrule
\multirow{5}{*}{ISRUC-S1}
   & Wake & 92.7 & 91.3 & 97.9 & 20098  \\
   & N1   & 74.4 & 61.9 & 97.4 & 11062  \\
   & N2   & 81.3 & 86.9 & 90.1 & 27511  \\
   & N3   & 87.2 & 87.8 & 96.8 & 17251  \\
   & REM  & 91.9 & 89.6 & 98.9 & 11265  \\
\midrule
\multirow{5}{*}{ISRUC-S3}
   & Wake & 85.4 & 90.3 & 96.1 & 1651   \\
   & N1   & 73.0 & 64.5 & 96.5 & 1215   \\
   & N2   & 89.9 & 89.9 & 95.6 & 2609   \\
   & N3   & 86.2 & 86.7 & 96.0 & 1914   \\
   & REM  & 89.3 & 90.4 & 98.3 & 1160   \\
\midrule
\multirow{5}{*}{Sleepedf-20}
   & Wake & 93.6 & 93.7 & 98.4 & 8285   \\
   & N1   & 65.9 & 65.0 & 97.6 & 2804   \\
   & N2   & 90.9 & 90.9 & 93.4 & 17799  \\
   & N3   & 85.4 & 85.5 & 97.7 & 5703   \\
   & REM  & 94.5 & 94.6 & 98.8 & 7717   \\
\midrule
\multirow{5}{*}{Sleepedf-78}
   & Wake & 94.5 & 93.7 & 97.3 & 63802  \\
   & N1   & 66.4 & 67.1 & 95.7 & 21229  \\
   & N2   & 90.3 & 90.8 & 94.6 & 68645  \\
   & N3   & 82.8 & 80.5 & 98.8 & 12883  \\
   & REM  & 91.6 & 92.8 & 98.7 & 25651  \\
\midrule
\multirow{5}{*}{SHHS}
   & Wake & 92.6 & 91.6 & 98.8 & 46319  \\
   & N1   & 57.8 & 59.3 & 98.5 & 10304  \\
   & N2   & 91.6 & 90.9 & 93.6 & 142125 \\
   & N3   & 89.3 & 89.6 & 97.6 & 60153  \\
   & REM  & 90.1 & 91.7 & 97.4 & 65953  \\
\bottomrule
\end{tabular}
\end{table}

\subsection{Results}
Tab.~\ref{tab:3} presents a comprehensive comparison between LGFNet and state-of-the-art methods across five public datasets.In cross-dataset comparisons, we highlight the superiority of LGFNet along two key dimensions: overall performance and robustness on difficult classes. The detailed confusion matrices for all datasets are shown in Fig.\ref{fig:4}, indicating that LGFNet achieves more balanced predictions across sleep stages, particularly improving the recognition of N1 and REM.

Per-stage performance (e.g., precision, F1-score, and specificity) is reported in Tab.~\ref{tab:4}. On the smaller yet more heterogeneous ISRUC dataset, our method achieves overall accuracies of 86.1\% and 85.7\%, macro F1 scores of 84.4\% and 83.5\%, and Cohen’s $\kappa$ values of 0.822 and 0.814 on the S3 and S1 splits, respectively. These results surpass the strongest existing single-channel baselines (e.g., MVF-SleepNet~[26] and SleepAC~[56]) by approximately 2–3.5\% in accuracy, 2–3.5\% in macro F1, and 0.03–0.04 in $\kappa$. More importantly, our model consistently achieves state-of-the-art performance on the most challenging N1 stage (64.5\% on S3 and 61.9\% on S1) while maintaining high recall on the REM stage (90.4\% on S3 and 89.6\% on S1). These results indicate that the observed performance gains are not obtained at the expense of difficult classes, but instead reflect improved discrimination across all sleep stages.

On the widely used Sleep-EDF benchmark, our method achieves an accuracy of 89.7\%, a macro F1 score of 85.9\%, and a Cohen’s $\kappa$ of 0.859 under 20-fold subject-wise cross-validation, and 88.7\%, 85.0\%, and 0.845, respectively, under the more rigorous 78-fold protocol. These results outperform recent strong competitors such as SleepViTransformer~[30], FlexibleSleepNet~[55], and the Multi-channel Transformer~[57]. Notably, LGFNet attains N1-stage recall rates of 65.0\% and 67.1\%, substantially exceeding the typical 49–57\% range reported in prior studies, while maintaining strong performance on the REM stage, with recall values of 94.6\% and 92.8\%, respectively.Fig. ~\ref{fig:5} presents a qualitative visualization of a representative subject, illustrating the temporal consistency of the predicted sleep stages and the associated confidence behavior of LGFNet.

When scaling to the large-scale and highly diverse SHHS dataset, our method continues to deliver strong performance, achieving an accuracy of 89.87\%, a macro F1 score of 84.6\%, and a Cohen’s $\kappa$ of 0.858, while further improving N1-stage recall to 59.3\%, which is notably higher than the 40–51\% range reported by most prior approaches. These results highlight the model’s strong transferability and robustness across markedly different population sizes, recording conditions, and data distributions.

The observed pattern of ``higher overall performance with markedly more stable difficult-class behavior'' is closely aligned with our core architectural contributions. The dual-branch design of LGFM-Encoder effectively decouples long-range dependencies from local time--frequency patterns, while the CTC objective encourages sharper and more temporally coherent emission distributions during training. In addition, the physiologically informed Viterbi decoding applied at inference imposes globally coherent trajectories by leveraging plausible stage-transition priors. Collectively, these mechanisms lead to consistent improvements in accuracy, macro~F1, and $\kappa$, with particularly significant gains on boundary-critical stages such as N1 and REM. Together, they represent a substantive step beyond conventional convolutional/recurrent hybrids and pure Transformer-based architectures, advancing both predictive accuracy and clinical reliability.


\begin{table*}[t]
\centering
\caption{Ablation study on key model components}
\label{tab:5}
\small 
\setlength{\tabcolsep}{8pt} 
\begin{tabular}{@{}lccc|ccc|ccc@{}}
\toprule
\multirow{2}{*}{\textbf{Baseline}} 
 & \multirow{2}{*}{\textbf{LFM}} 
 & \multirow{2}{*}{\textbf{GFM}} 
 & \multirow{2}{*}{\textbf{CTC loss}} 
 & \multicolumn{3}{c|}{\textbf{Sleep EDF-78}} 
 & \multicolumn{3}{c}{\textbf{Sleep EDF-20}} \\
\cmidrule(lr{0.5em}){5-7} \cmidrule(l{0.5em}){8-10}
 &   &   &   & Acc      & F1       & Kappa   & Acc      & F1       & Kappa    \\
\midrule
\checkmark & \checkmark & × & \checkmark & 80.4\%   & 73.4\%   & 0.733   & 79.5\%   & 74.1\%   & 0.720 \\
\checkmark & \checkmark & \checkmark & × & 84.9\%   & 80.1\%   & 0.793   & 84.4\%   & 79.7\%   & 0.787 \\
\checkmark & × & × & \checkmark & 70.8\%   & 63.4\%   & 0.603   & 71.7\%   & 65.3\%   & 0.614 \\
\checkmark & × & \checkmark & \checkmark & 81.5\%   & 76.1\%   & 0.747   & 79.8\%   & 73.3\%   & 0.724 \\
\checkmark & \checkmark & \checkmark & \checkmark & \textbf{88.7\%} & \textbf{85.0\%} & \textbf{0.845} 
                     & \textbf{89.7\%} & \textbf{85.9\%} & \textbf{0.859} \\
\bottomrule
\setlength{\textfloatsep}{6pt}
\setlength{\intextsep}{6pt}
\end{tabular}
\end{table*}

\begin{table*}[t]
\centering
\caption{Ablation study on decoder components}
\label{tab:6}
\small
\setlength{\tabcolsep}{10pt}
\begin{tabular}{@{}lccc|ccc|ccc@{}}
\toprule
\multirow{2}{*}{\textbf{Baseline}} 
 & \multirow{2}{*}{\textbf{CTC}} 
 & \multirow{2}{*}{\textbf{Viterbi}} 
 & \multirow{2}{*}{\textbf{}} 
 & \multicolumn{3}{c|}{\textbf{Sleep EDF-78}} 
 & \multicolumn{3}{c}{\textbf{Sleep EDF-20}} \\
\cmidrule(lr{0.5em}){5-7} \cmidrule(l{0.5em}){8-10}
 &  &  &  & Acc      & F1       & Kappa    & Acc      & F1       & Kappa     \\
\midrule
\checkmark & × & × &  & 82.6\%   & 77.7\%   & 0.763    & 80.0\%   & 74.0\%   & 0.727  \\
\checkmark & \checkmark & × &  & 86.5\%   & 82.4\%   & 0.815    & 84.0\%   & 78.6\%   & 0.789  \\
\checkmark & × & \checkmark &  & 87.2\%   & 83.2\%   & 0.825   & 83.6\%   & 77.9\%   & 0.776  \\
\checkmark & \checkmark & \checkmark &  & \textbf{88.7\%} & \textbf{85.0\%} & \textbf{0.845} 
                                  & \textbf{89.7\%} & \textbf{85.9\%} & \textbf{0.859} \\
\bottomrule
\setlength{\textfloatsep}{6pt}
\setlength{\intextsep}{6pt}

\end{tabular}
\end{table*}
\captionsetup{belowskip=2pt}

\begin{figure*}[!t]
\addtocounter{figure}{+2}
\centering
\includegraphics[width=0.98\textwidth]{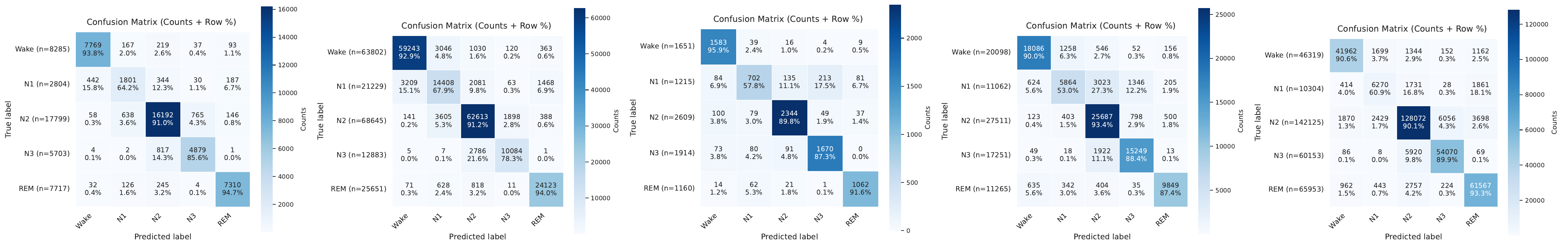}
\includegraphics[width=0.98\textwidth]{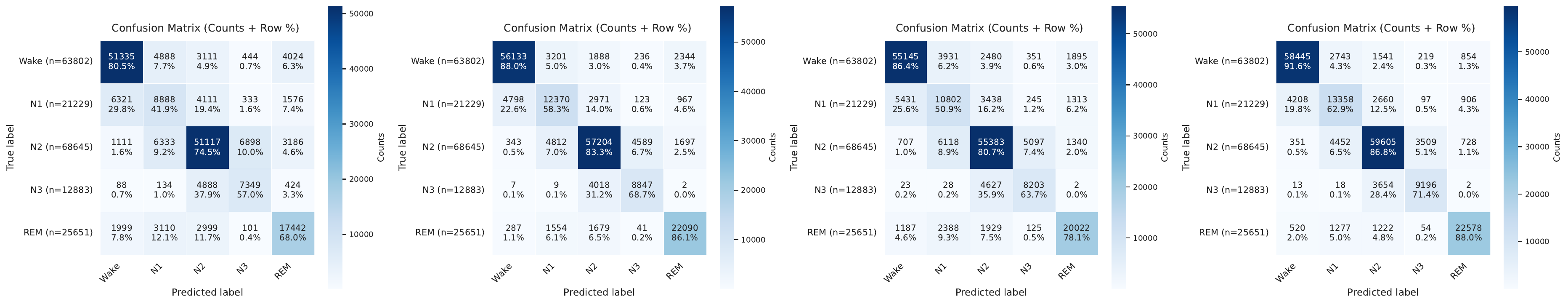}
\includegraphics[width=0.98\textwidth]{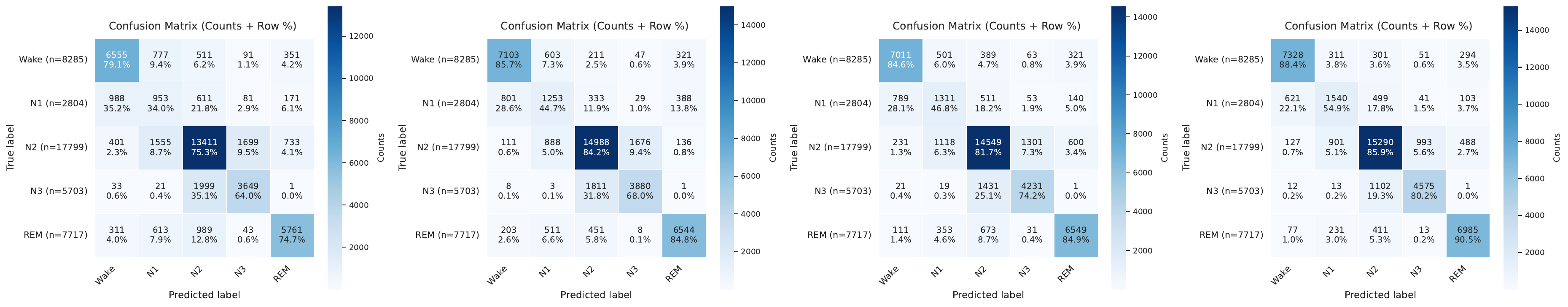}
\captionsetup{font=scriptsize}
\caption{%
The first row shows the confusion matrices of LGFNet on Sleep-EDF 20/78, ISRUC S3/S1, and SHHS; the second and third rows show the confusion matrices from the ablation study for GFM-only, LFM-only, and LGFNet without CTC-guided decoding.
}
    
\label{fig:4}
\end{figure*}

\begin{figure*}[!t]
\centering
\includegraphics[width=0.98\textwidth]{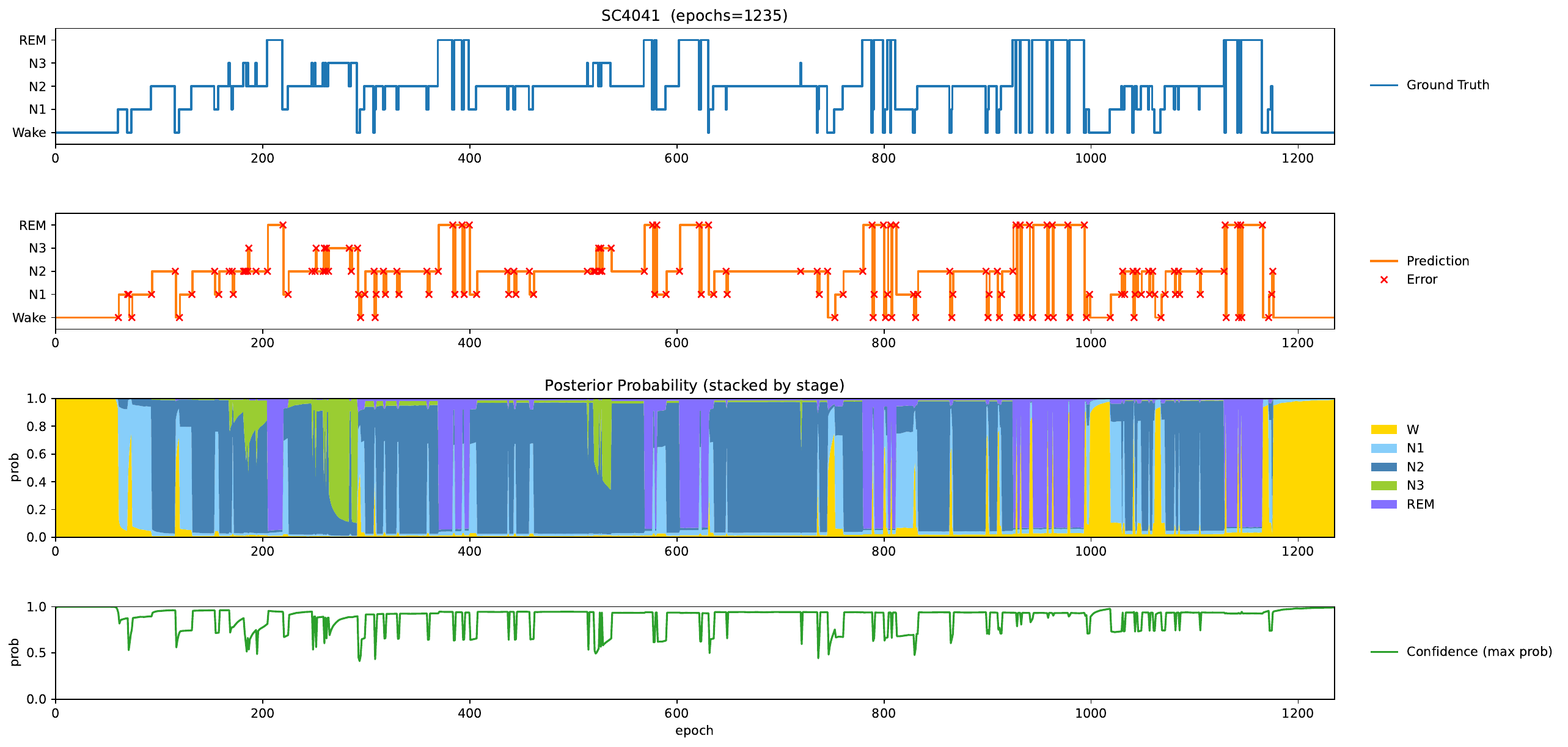}
\captionsetup{font=scriptsize}
\caption{%
Visualization of the estimated confidence for SC4041 of SleepEDF-78.
The first line is output hypnogram produced by the proposed LGFNet for Subject~1 of the SleepEDF dataset, compared to the ground-truth hypnogram
The second line is errors are indicated by red markers.
The third line is confidence score defined as the maximum posterior among the five stages.
The last line is posterior probability distribution over sleep stages. 
}
    
\label{fig:5}
\end{figure*}

\begin{figure*}[!t]
    \centering

    \begin{subfigure}[t]{0.48\textwidth}
        \centering
        \includegraphics[width=\linewidth]{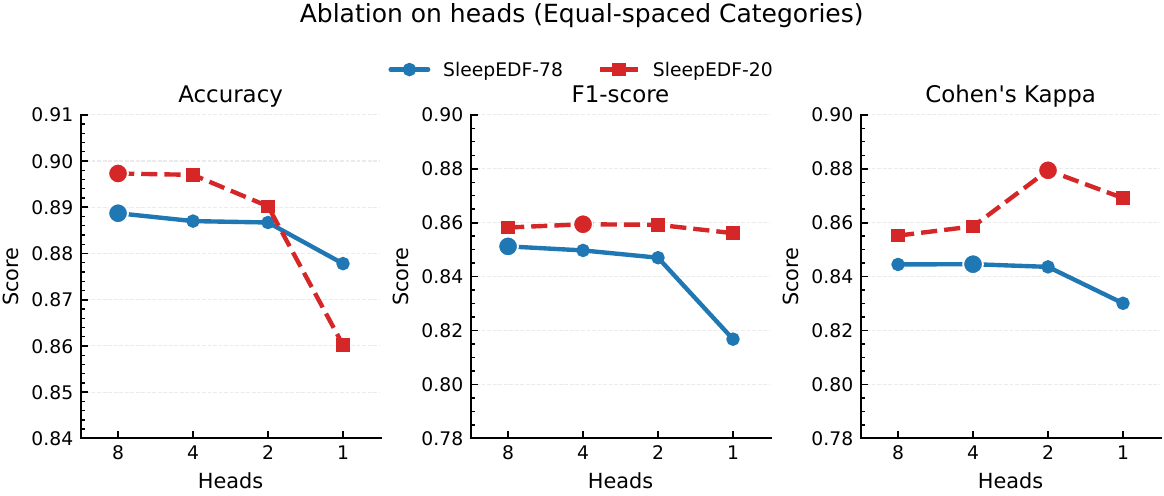}
        \caption{\textbf{Figure 6:} Changes in accuracy, Cohen’s kappa, and macro-F1 across different numbers of attention heads.}
        \label{fig:6}
    \end{subfigure}
    \hfill
    \begin{subfigure}[t]{0.48\textwidth}
        \centering
        \includegraphics[width=\linewidth]{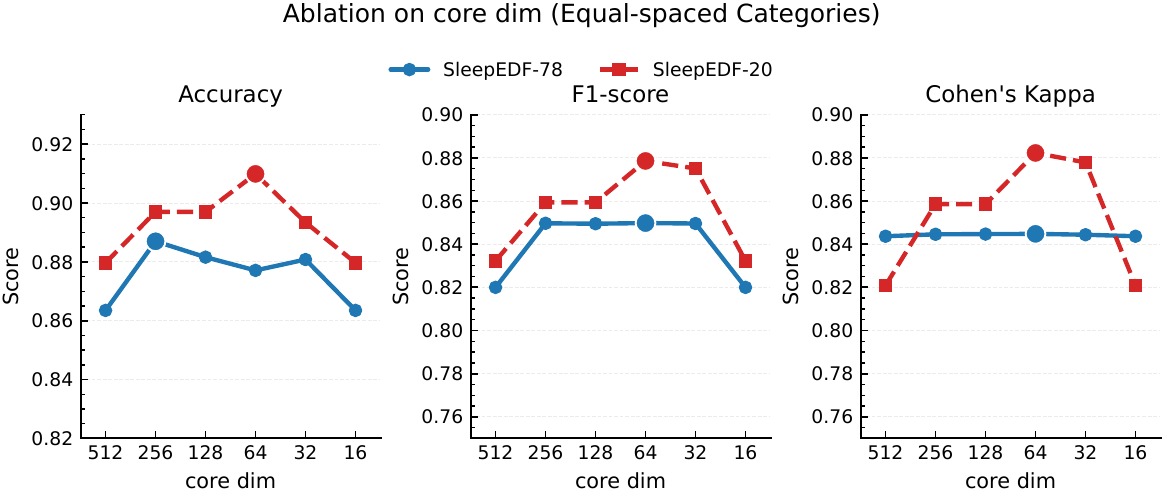}
        \caption{\textbf{Figure 7:} Changes in accuracy, Cohen’s kappa, and macro-F1 across different core dimensions.}
        \label{fig:7}
    \end{subfigure}

    \vspace{4pt}

    \begin{subfigure}[t]{0.98\textwidth}
        \centering
        \includegraphics[width=\linewidth]{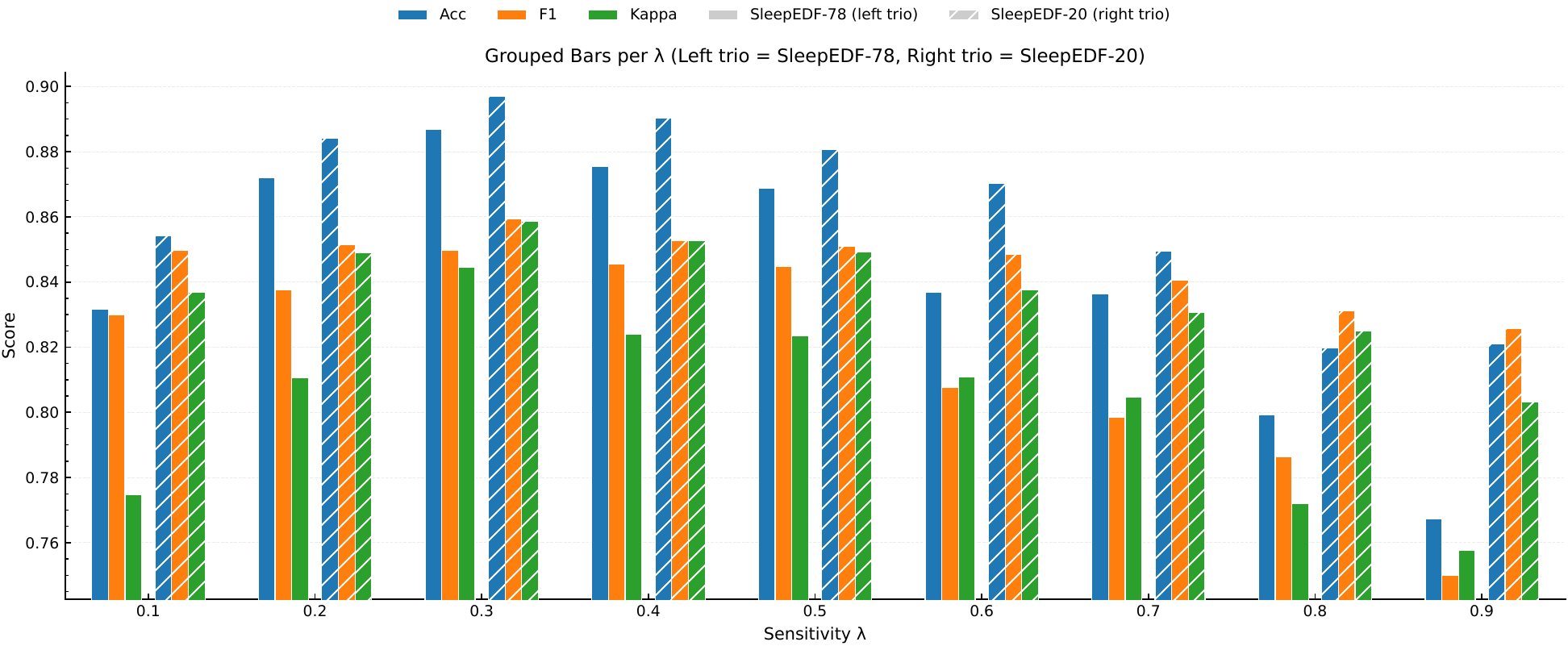}\vspace{-0.8pt}
        \includegraphics[width=\linewidth]{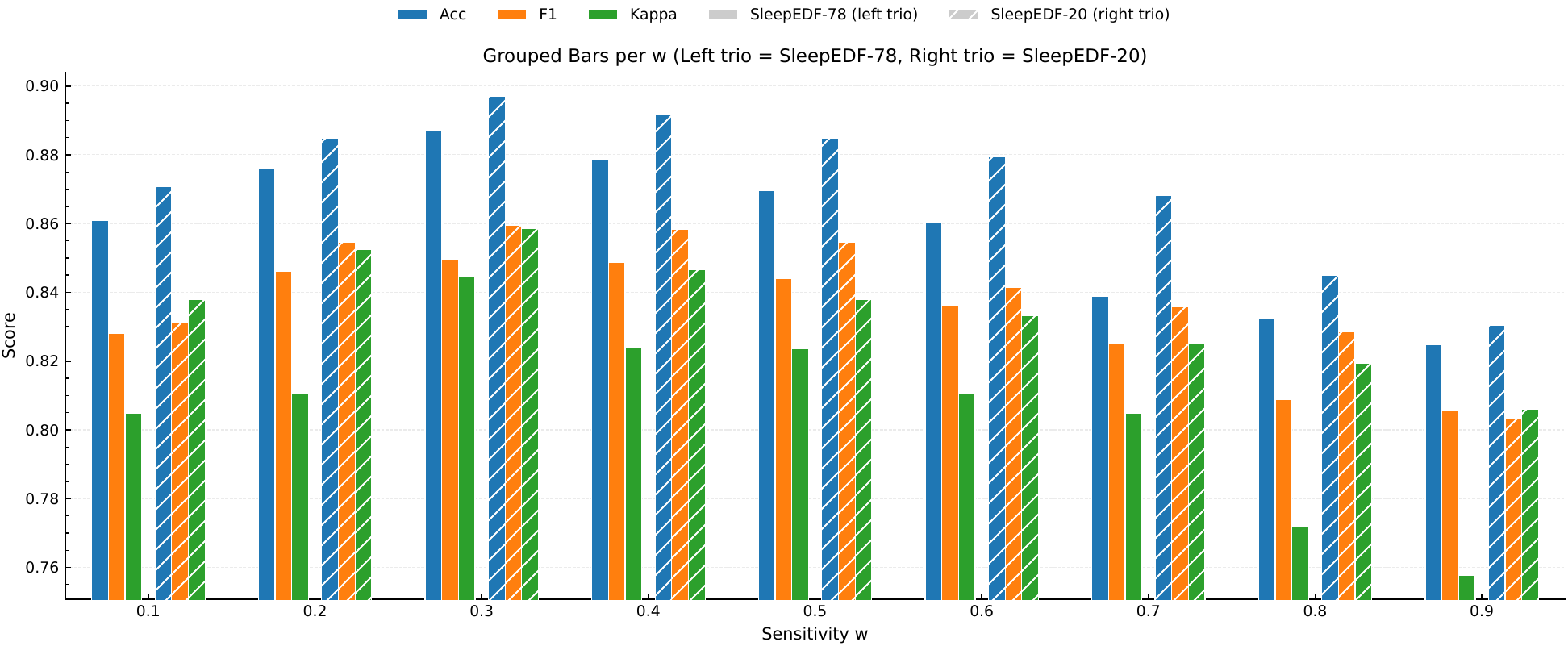}
        \caption{\textbf{Figure 8:} Changes in accuracy, Cohen’s kappa, and macro-F1 across the multi-task weight $\lambda$ and transition weight $w$.}
        \label{fig:8}
    \end{subfigure}
\end{figure*}

\section{Ablation Studies}
\subsection{Module Ablation}
\label{sec:ablation}

We conduct controlled ablation studies starting from a deliberately minimal baseline to quantify the contribution of each model component. The baseline is a vanilla Transformer sequence-to-sequence model without CTC, in which both training and inference rely solely on cross-entropy loss with standard autoregressive decoding. The ablation results of key components in LGFNet are summarized in Tab.~\ref{tab:5}. In this setting, the LGFM-Encoder is simplified to a single-branch vanilla Transformer encoder paired with a standard Transformer decoder. To isolate the effect of incorporating CTC, we restore the full LGFM-Encoder and replace CE-only training with joint CTC and cross-entropy optimization, together with CTC-guided decoding. On the Sleep-EDF datasets under the 78- and 20-fold protocols, performance improves from 84.4–84.9\% accuracy, 79.7–80.1\% macro F1, and $\kappa$ values of 0.787–0.793 to 88.7–89.7\% accuracy, 85.0–85.9\% macro F1, and $\kappa$ values of 0.845–0.859. This corresponds to gains of 3.8–5.3\% in accuracy and 4.9–6.2\% in macro F1.

Building on the CTC-enabled setting, we further ablate the two branches of the LGFM-Encoder. Removing the GFM (global fusion via multi-head self-attention) while retaining only the LFM (local gated MLP) leads to accuracy drops of 8–10\%, macro F1 reductions of around 12\%, and decreases of 11–14\% in Cohen’s $\kappa$. Conversely, removing the LFM while keeping only the GFM results in decreases of 7–10\% in accuracy, 9–13\% in macro F1, and 10–14\% in $\kappa$. Eliminating both branches, leaving only a vanilla backbone equipped with a CTC head, further reduces accuracy to approximately 71\%. These results collectively indicate that CTC improves temporal alignment, GFM captures long-range dependencies, and LFM models fine-grained local dynamics; only their synergistic integration yields the optimal LGFNet configuration.

\subsection{Ablation on CTC and Viterbi}
We start from a restrained baseline consisting of a vanilla Transformer encoder and a standard Transformer decoder, trained and decoded solely with cross-entropy, without the use of CTC. Tab.~\ref{tab:6} quantifies the contributions of CTC training and Viterbi decoding to the overall performance.On the Sleep-EDF dataset under the 78-fold and 20-fold evaluation protocols, this baseline achieves accuracies of 80.0–82.6\%, macro F1 scores of 74.0–77.7\%, and Cohen’s $\kappa$ values of 0.727–0.763.

Introducing CTC during training while using the full LGFM-Encoder, but still decoding without Viterbi, increases performance to 84.0–86.5\% accuracy, 78.6–82.4\% macro F1, and Cohen’s $\kappa$ values of 0.789–0.815. These results correspond to absolute improvements of 3.8–4.1\% in accuracy, 4.6–4.7\% in macro F1, and 5.2–6.2\% in $\kappa$ over the baseline, indicating that CTC alone, through alignment learning and path compression, stabilizes frame-level emissions and sharpens decision boundaries.

Conversely, training with cross-entropy only while applying Viterbi decoding at inference, using physiological transition priors, yields improvements of 3.6–4.6\% in accuracy, 3.9–5.5\% in macro F1, and 4.9–6.2\% in Cohen’s $\kappa$ relative to the baseline. Across evaluation folds, this “Viterbi-only” setting sometimes slightly surpasses and sometimes slightly underperforms the “CTC-only” configuration, suggesting that Viterbi primarily provides global sequence-level correction at test time, whereas CTC more fundamentally enhances alignment learning during training.

Enabling both CTC during training and Viterbi decoding at inference yields a clear synergistic effect, achieving accuracies of 88.7–89.7\%, macro F1 scores of 85.0–85.9\%, and Cohen’s $\kappa$ values of 0.845–0.859. Relative to the CTC-only configuration, this corresponds to additional gains of 2.2–5.7\% in accuracy, 2.5–7.3\% in macro F1, and 3.0–7.0\% in $\kappa$. Compared to the minimal baseline, the cumulative improvements reach 6.1–9.7\% in accuracy, 7.3–11.9\% in macro F1, and 8.2–13.2\% in $\kappa$. These results validate the design choices: CTC provides essential temporal alignment and path constraints that stabilize frame-level emissions, while Viterbi decoding enforces global Markov priors at inference to produce coherent stage trajectories. In combination, these components drive the substantial performance gains observed in the full system.

\subsection{Super Parameter Sensitive}
\label{sec:sensitivity}

We evaluate three critical hyperparameter groups and show that the default configuration is both effective and robust. First, we vary the Viterbi fusion weight $\alpha_{\text{fuse}}$, which balances decoder and CTC emissions at inference. As $\alpha_{\text{fuse}}$ increases from 0.1 to 0.3, performance improves steadily and peaks at $\alpha_{\text{fuse}}=0.3$, achieving accuracies of 89.7\% and 88.7\%, macro F1 scores of 85.9\% and 84.97\%, and Cohen’s $\kappa$ values of 0.859 and 0.8456 on the two evaluation splits, respectively. Further increasing $\alpha_{\text{fuse}}$ from 0.4 to 0.9 leads to an inverted-U-shaped decline in performance, indicating that over-reliance on either emission source degrades the fused decoding path. Accordingly, we adopt $\alpha_{\text{fuse}}=0.3$ as the default, which remains stable within the range $[0.2, 0.4]$.

Second, we sweep the multi-task weight $\lambda$, which balances the CTC and cross-entropy objectives. A similar inverted-U-shaped pattern is observed, with optimal performance at $\lambda=0.3$, consistent with the peak identified above. Noticeable degradation occurs for $\lambda \le 0.1$ or $\lambda \ge 0.6$, indicating that either under-weighting or over-weighting the CTC objective is detrimental. This setting consistently yields the best performance and the most stable convergence across both evaluation folds.

Finally, we examine model capacity. Increasing the number of attention heads from 1 to 4 (and further to 8) improves accuracy from 86.0–87.8\% to 88.7–89.7\%, confirming the benefit of multi-head subspace decomposition.As shown in Fig. 6, increasing the number of attention heads from 1 to 4 consistently improves performance, while further increasing to 8 brings marginal gains. The performance gap between 4 and 8 heads is small, while 4 heads are more computationally efficient; therefore, we adopt 4 heads by default. Fig. 7 further shows that a core dimension of 256 yields the most stable and competitive results across datasets.With respect to hidden dimensionality, a model size of $d_{\text{model}}=256$ provides the most balanced results across folds, achieving accuracies of 88.7–89.7\%, macro F1 scores of 84.97–85.94\%, and Cohen’s $\kappa$ values of 0.845–0.859. Smaller dimensions (64 and 128) can occasionally match individual metrics but exhibit higher variance, whereas a larger size of 512 tends to overfit. Very small dimensions (16 and 32) consistently underfit.

In summary,the sensitivity analysis with respect to the multi-task weight $\lambda$ and transition-related weight w is summarized in Fig. 8, where an inverted U-shaped trend can be observed,and the default configuration ($\alpha_{\text{fuse}}=0.3$, $\lambda=0.3$, $n_{\text{heads}}=4$, and $d_{\text{model}}=256$) consistently attains near-optimal performance across random splits and aligns well with the overall design rationale. Specifically, CTC and the autoregressive decoder play complementary roles, while a moderate number of attention heads combined with an appropriate hidden dimensionality provides sufficient representational capacity without introducing unnecessary model complexity.

\begin{table}[H] 
\centering
\caption{Comparison of Acc, Params, MTT (per 100 epochs), and MU.}
\setlength{\tabcolsep}{1.0pt}
\label{tab:7}
\begin{tabular}{lcccc}
\toprule
Methods & Acc & Params & MTT & MU \\
\midrule
FlexibleSleepNet           & 87.0  & 27.5M & 68.5min & 125.1MB \\
FlexibleSleepNet-Small     & 86.5  & 14.3M & 31.7min & 65.2MB  \\
FlexibleSleepNet-Large     & 87.1 & 49.4M & 72.0min & 218.6MB \\
Multi-channel              & 85.0  & 27.2M & 2.5hr   & 125.0MB \\
SleepTransformer           & 84.9  & 3.7M  & 2.3hr   & --      \\
DeepSleepNet               & 77.8  & 23.0M & 7.2hr   & --      \\
\midrule
\textbf{Ours}              & \textbf{88.7} & \textbf{16.6M} & \textbf{1.7hr} & \textbf{63.52MB} \\
\bottomrule
\end{tabular}
\end{table}

\section{Discussion}
This work presents an end-to-end framework for single-channel EEG sleep staging that integrates learnable emission modeling with physiologically grounded sequence constraints. By jointly training a cross-entropy decoder and a CTC alignment head and fusing their emissions at inference using Viterbi decoding with transition priors, the proposed design mitigates exposure bias and scale mismatch between emission streams, leading to improved temporal consistency and fine-grained discrimination, particularly for challenging stages such as N1 and N3.
The method achieves stable and competitive performance across multiple public datasets with diverse populations, recording protocols, and sampling rates, while maintaining favorable efficiency in terms of parameter count, memory footprint, and training cost. A comparison of model efficiency (e.g., parameter count, memory footprint, and training cost) is provided in Tab.~\ref{tab:7}. Ablation and sensitivity analyses further confirm the complementary roles of local and global modeling branches, as well as the independent and additive contributions of CTC and Viterbi decoding, with key hyperparameters showing robust behavior across settings. Overall, the results suggest that incorporating lightweight sequence constraints can be more effective than increasing model capacity alone for practical single-channel sleep staging.

Despite these advantages, several limitations remain. The current framework is restricted to single-channel EEG, and its extension to multi-lead or multimodal settings has not yet been validated. In addition, the CTC upsampling strategy relies on an approximate alignment, and the use of empirically estimated transition priors may require recalibration across populations, devices, and preprocessing pipelines. Robustness under severe noise conditions and pathological sleep patterns also warrants further study. Future work will explore multi-lead and multimodal fusion, self-supervised pretraining to further improve difficult stages such as N1, and the incorporation of learnable front-end representations and neural transition models to reduce reliance on handcrafted priors. We also plan to investigate cross-device adaptation and uncertainty calibration to enhance clinical transferability, together with releasing code and pretrained models to support reproducibility and community benchmarking.

\end{document}